\documentclass[runningheads]{llncs}
\usepackage{graphicx}
\usepackage{cite}
\usepackage[pdfa]{hyperref}
\hypersetup{
colorlinks=true,
filecolor=blue,
urlcolor=blue
}

\usepackage{amsmath}
\newcommand{\XX}{\mathbf{X}}
\newcommand{\UU}{\mathbf{U}}
\newcommand{\YY}{\mathbf{Y}}
\newcommand{\FF}{\mathbf{F}}
\begin{document}
\title{Matrix Lie Maps and Polynomial Neural Networks for Solving Differential Equations}
\titlerunning{Matrix Lie Maps and Neural Networks for Solving Differential Equations}
\author{Andrei Ivanov\inst{1}\orcidID{0000-0002-1663-3721} \and
Sergei Andrianov\inst{1}\orcidID{0000-0002-4648-2047}}
\authorrunning{A. Ivanov, S. Andrianov}
\institute{Saint Petersburg State University, Saint Petersburg, Russia\\
%10969257 Canada Inc., Toronto, Canada\\
\email{05x.andrey@gmail.com}}

\maketitle              % typeset the header of the contribution
\begin{abstract}
The coincidence between polynomial neural networks and matrix Lie maps is discussed in the article. The matrix form of Lie transform is an approximation of the general solution of the nonlinear system of ordinary differential equations. It can be used for solving systems of differential equations more efficiently than traditional step-by-step numerical methods. Implementation of the Lie map as a polynomial neural network provides a tool for both simulation and data-driven identification of dynamical systems. If the differential equation is provided, training a neural network is unnecessary. The weights of the network can be directly calculated from the equation. On the other hand, for data-driven system learning, the weights can be fitted without any assumptions in view of differential equations. The proposed technique is discussed in the examples of both ordinary and partial differential equations. The building of a polynomial neural network that simulates the Van der Pol oscillator is discussed. For this example, we consider learning the dynamics from a single solution of the system. We also demonstrate the building of the neural network that describes the solution of Burgers' equation that is a fundamental partial differential equation.

\keywords{Polynomial neural networks  \and Matrix Lie maps \and Differential equations.}
\end{abstract}
\section{Introduction}
Traditional methods for solving systems of differential equations imply a numerical step-by-step integration of the system. For some problems, this integration leads to time-consuming algorithms because of the limitations on the time interval that is used to achieve the necessary accuracy of the solution. From this perspective, neural networks as universal function approximation can be applied for the construction of the solution in a more performance way.

In the article \cite{ref1}, the method to solve initial and boundary value problems using feedforward neural networks is proposed. The solution of the differential equation is written as a sum of two parts. The first part satisfies the initial/boundary conditions. The second part corresponds to a neural network output. The same technique is applied for solving Stokes problem in \cite{ref2,ref3} and implemented in code in \cite{ref4}.

In the article \cite{ref5}, the neural network is trained to satisfy the differential operator, initial condition, and boundary conditions for the partial differential equation (PDE). The authors in \cite{ref6} translate a PDE to a stochastic control problem and use deep reinforcement learning for an approximation of derivative of the solution with respect to the space coordinate.

Other approaches rely on the implementation of a traditional step-by-step integrating method in a neural network basis \cite{ref7,ref8}. In the article \cite{ref8}, the author proposes such an architecture. After fitting, the neural network produces an optimal finite difference scheme for a specific system. The backpropagation technique through an ordinary differential equation (ODE) solver is proposed in \cite{ref9}. The authors construct a certain type of neural network that is analogous to a discretized differential equation. This group of methods requires a traditional numerical method to simulate dynamics.

Polynomial neural networks are also widely presented in the literature \cite{ref10,ref11,ref12}. In the article \cite{ref10}, the polynomial architecture that approximates differential equations is proposed. The Legendre polynomial is chosen as a basis function of hidden neurons in \cite{ref11}. In these articles, the polynomial architectures are used as black box models, and the authors do not explain its connection to the theory of ODEs.

In all the described approaches, the neural networks are trained to consider the initial conditions of the differential equations. This means that the neural network should be trained each time when the initial conditions are changed. The above-described techniques are applicable to the general form of differential equations but are able to provide only a particular solution of the system.

In the article, we consider polynomial differential equations. Such nonlinear systems arise in different fields such as automated control, robotics, mechanical and biological systems, chemical reactions, drug development, molecular dynamics, and so on. Moreover, often it is possible to transform a nonlinear equation to a polynomial view with some level of accuracy.

For polynomial differential equations, it is possible to build a polynomial neural network that is based on the matrix Lie transform and approximates the general solution of the system of equations. Having a Lie transform--based neural network for such a system, dynamics for different initial conditions can be estimated without refitting of the neural network. Additionally, we completely avoid numerical ODE solvers in both simulation and data-driven system learning by describing the dynamics with maps instead of step-by-step integrating.

\section{Proposed Neural Network}
\label{sec2}
The proposed architecture is a neural network representation of a Lie propagator for dynamical systems integration that is introduced in \cite{ref13} and is commonly used in the charged particle dynamics simulation \cite{ref13,ref14}. We consider dynamical systems that can be described by nonlinear ordinary differential equations, 
\begin{equation}
\label{odesystem}
\frac{d}{dt}\XX = \FF(t, \XX) = \sum_{k=0}^{\infty} P^{1k}(t)\XX^{[k]},
\end{equation}
where $t$ is an independent variable, $\XX \in R^n$ is a state vector, and $\XX^{[k]}$ means $k$-th Kronecker power of vector $\XX$. There is an assumption that function $\FF$ can be expanded in Taylor series with respect to the components of $\XX^{[k]}$.

The solution of (\ref{odesystem}) in its convergence region can be presented in the series \cite{ref15,ref16},
\begin{equation}
\label{Lie}
\XX(t|t_0) =  \mathcal M(t|t_0) \circ \XX_0 =  \sum_{k=0}^{\infty} M^{1k}(t|t_0)\XX_0^{[k]},
\end{equation}
where $\XX_0 = \XX(t_0)$ In \cite{ref14}, it is shown how to calculate matrices $M^{1k}$ by introducing new matrices $P^{ij}$.
%$P^{ij} = P^{1(j-i+1)}P^{(i-1)(j-1)}$.
The main idea is replacing 
%a differential equation
(\ref{odesystem}) by the equation
\begin{equation}
\label{Lie_map}
\frac{d}{dt} M^{ik}(t|t_0) = \sum_{j=i}^{k} P^{ij}(t)M^{jk}(t|t_0),\;1\leq i < k.
\end{equation}
%where $P^{ij} = P^{1(j-i+1)}P^{(i-1)(j-1)}$.
This equation should be solved with initial condition $M^{kk}(t_0) = I^{[k]},\;M^{jk}(t_0) = 0, j\neq k$, where $I$ is the identity matrix. Theoretical estimations of accuracy and convergence of the truncated series in solving of ODEs can be found in \cite{ref17}. 

%In this article we do not consider these questions and truncate the series (\ref{Lie}) based on experimentation.

%In this way, one has two equivalent representations of dynamical systems.
The transformation $\mathcal{M}$ can be considered as a discrete approximation of the evolution operator of (\ref{odesystem}) for initial time $t_0$ and interval $\Delta t$. This means that the evolution of the state vector $\XX_0 = \XX(t_0)$ during time $\Delta t$ can be approximately calculated as $\YY = \mathcal{M} \circ \XX_0$. Hence, instead of solving the system of ODEs numerically, one can apply a calculated map and avoid a step-by-step integrating.
%The transformation $\mathcal{M}$ can be considered as a discrete approximation of the evolution operator of (\ref{odesystem}) for predefined initial time $t_0$ and time interval $\Delta t$. This means that the evolution of initial state vector $\XX_0 = \XX(t_0)$ during time $\Delta t$ can be approximately calculated as $\YY = \XX(t_0, \XX_0, \Delta t,) = \mathcal{M} \circ \XX_0$. Hence, instead of solving the system of ODEs numerically, one can apply a calculated map and avoid a step-by-step integrating of the ODEs.

\subsection{Neural Network Representation of Matrix Lie Transform}
The proposed neural network implements map $\mathcal M : \XX\rightarrow \YY$ in form of
%using the following polynomial transformation,
\begin{equation}
	\label{nnmap}
	\YY = W_0 + W_1\,\XX+W_2\,\XX^{[2]}+\ldots+W_k\,\XX^{[k]},
\end{equation}
where $\XX, \YY \in R^n$, $W_i$ are weight matrices, and $\XX^{[k]}$ means $k$-th the Kronecker power of vector $\XX$. For a given system of ODEs (\ref{odesystem}), one can compute matrices $W_i = M^{1k}$ in accordance with (\ref{Lie_map}) up to the necessary order of nonlinearity. 

\begin{figure}
\centering
\includegraphics[width=0.55\textwidth]{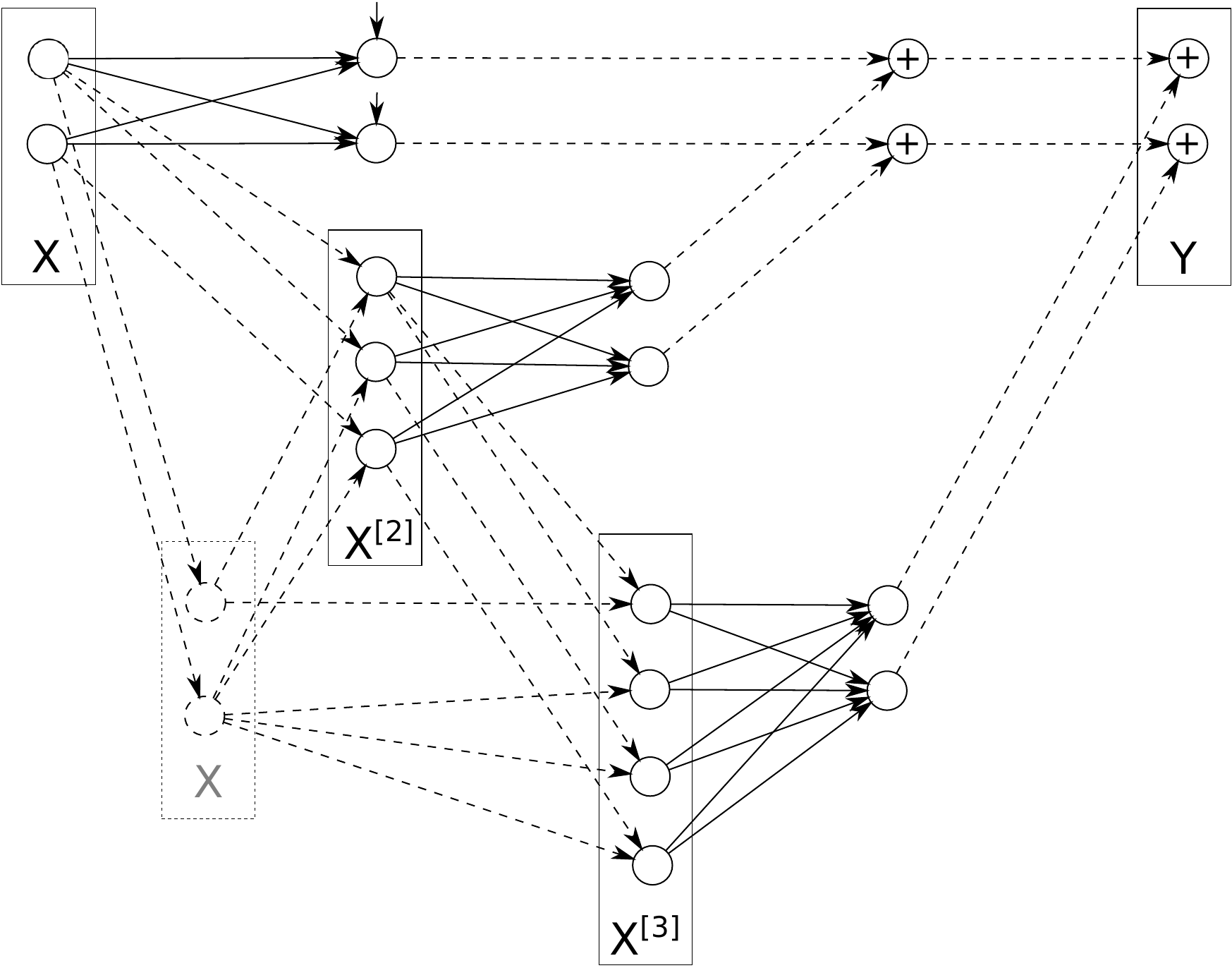}
\caption{Neural network representation of third order matrix Lie map.} \label{fig1}
\end{figure}

Fig.~\ref{fig1} presents a neural network for map (\ref{nnmap}) up to the third order of nonlinearities for a two-dimensional state. In each layer, the input vector $\XX = (x_1, x_2)$ is consequently transformed into $\XX^{[2]} = (x_1^2, x_1x_2, x_2^2)$ and $\XX^{[3]} = (x_1^3, x_1^2x_2, x_1x_2^2$, $x_2^3)$ where weighted sum is applied. The output Y equals to the sum of results from every layer. In the example, we reduce Kronecker powers for decreasing of weights matrices dimension (e.g., $\XX^{[2]} = (x_1^2, x_1x_2, x_2x_1, x_2^2) \rightarrow (x_1^2, x_1x_2, x_2^2)$).

\subsection{Fitting Neural Network}
If the differential equation is provided, the training of neural network is not necessary. The weights of the network can be calculated directly from the equation following the relation (\ref{Lie_map}). On the other hand, for data-driven system learning, the weights in form of (\ref{nnmap}) can be fitted without any assumptions on view of differential equations. 

\noindent
To fit a proposed neural network, the training data is presented as a multivariate time series (table~\ref{table1}) that describes the evolution of the state vector of the dynamical system in a discrete time. In a general case, each step $t_i \rightarrow t_{i+1}$ should be described as map $\mathcal M_i(t_i): \XX_i \rightarrow \XX_{i+1}$, but if the system (\ref{odesystem}) is time independent, then weights $W_i$ depends only on time interval $\Delta t = t_{i+1} - t_i$.

\begin{table}
\centering
\caption{Discrete states of a dynamical system for training the proposed network.}\label{table1}
\begin{tabular}{lcccr}
\hline
$t_0$ & $t_1$ & $\ldots$ & $t_{m-1}$ & $t_m$ \\
\hline
$x_0(t_0)$ & $x_0(t_1)$ & $\ldots$ & $x_0(t_{m-1})$ & $x_0(t_m)$ \\
$x_1(t_0)$ & $x_1(t_1)$ & $\ldots$ & $x_1(t_{m-1})$& $x_m(t_m)$ \\
$\ldots$   &            & $\ldots$ & &            \\
$x_n(t_0)$ & $x_n(t_1)$ & $\ldots$ & $x_n(t_{m-1})$ & $x_n(t_m)$ \\
\hline
INPUT $\rightarrow$ & $\mathcal M_1$ $\rightarrow$ & $\ldots$ & $\rightarrow$ $\mathcal M_m$ $\rightarrow$ & OUTPUT \\
\hline
\end{tabular}
\end{table}

\section{Ordinary Differential Equations}
\label{sec3}
In this section, we consider the Van der Pol oscillator. The equation is widely used in the physical sciences and engineering and can be used for the description of the pneumatic hammer, steam engine, periodic occurrence of epidemics, economic crises, depressions, and heartbeat. The equation has well-studied dynamics and is widely used for testing of numerical methods (e.g., \cite{ref18}).

\subsection{Simulation of the Van der Pol Oscillator}
The Van der Pol oscillator is defined as the system of ODEs $x'' = x' - x - x^2x'$ that can be presented in the form of
\begin{equation}
\label{vdp_ode}
    \begin{aligned}
        x' &= y,\\
        y' &= y-x-x^2y.
    \end{aligned}
\end{equation}

\begin{figure}
\centering
\includegraphics[width=0.65\textwidth]{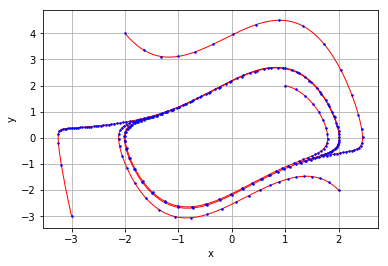}
\caption{Simulation of the Van der Pol oscillator. Red lines for the implicit Adams method of eighth order, blue dots for matrix Lie map of third order.}
\label{fig2}
\end{figure}

\noindent
The results of numerical integration of the system with the implicit Adams method of eighth order with the maximum time step $\Delta t = 0.01$ are presented in Fig. \ref{fig2} with red lines. The four different particular solutions with initial conditions, $(-2, 4), (1, 2), (2, -2)$, and $(-3, -3)$, were calculated.

\noindent
Another method for simulating the dynamics is mapping approach. The weights of the matrix Lie map can be calculated up to the necessary order of nonlinearity based on the equation (\ref{Lie_map}). For instance, for the third order and the same time interval, it yields weight matrices
\begin{equation*}
\begin{aligned}
&W_0 = 
\begin{pmatrix}
0\\
0\\
\end{pmatrix};\;
W_1 =
\begin{pmatrix}
0.99995067&  0.01004917\\
-0.01004917&  1.00999984\\
\end{pmatrix};\;
W_2 = 
\begin{pmatrix}
0&  0&  0\\
0&  0&  0\\
\end{pmatrix};\\
\\
&W_3=
\begin{pmatrix}
1.59504733e\text{-}7& -4.94822066e\text{-}5& -3.20576750e\text{-}7& -7.90629025e\text{-}10\\
4.94821629e\text{-}5& -1.00975145e\text{-}2& -9.96173322e\text{-}5& -3.30168067e\text{-}07\\
\end{pmatrix}.
\end{aligned}
\end{equation*}

\noindent
The corresponding polynomial neural network implements transformation
$$
\XX_{i+1} = \mathcal M\circ \XX_i =
W_0 + W_1
\begin{pmatrix}
x_i\\
y_i\\
\end{pmatrix}
+W_2
\begin{pmatrix}
x_i^2\\
x_iy_i\\
y_i^2
\end{pmatrix}
+W_3
\begin{pmatrix}
x_i^3\\
x_i^2y_i\\
x_iy_i^2\\
y_i^3
\end{pmatrix}.
$$

\noindent
The results of the numerical integration of the system with the neural network are presented in Fig. \ref{fig2} with blue dots.
Note that for the matrix Lie maps, the accuracy of the truncation of the series (order of nonlinearity of the transformation) and the accuracy of weights calculation should be considered separately. The theory of the accuracy and convergence of the truncated series (\ref{Lie}) in solving ODEs can be found in \cite{ref13,ref17}.

From a practical perspective, the accuracy of the simulation provided by a polynomial neural network can be estimated with respect to the traditional numerical solver. For example, the mean relative errors between the predictions of the Lie map--based networks of the third, fifth, and seventh orders of nonlinearity with respect to the numerical solution calculated with the Adams method of eighth order are equal to $0.0110$, $0.0004$, and $4.7\cdot10^{-6}$, respectively. 

\subsection{Learning of the Van der Pol Oscillator}
In the previous section, we described how weights for the proposed polynomial neural network can be calculated based on the equation. On the other hand, when the equation is not known, but a particular solution is provided, the weights can be fitted by a neural network without any assumptions in view of differential equations.

A particular solution $\{\XX_i\}_{i=1;n}$ of the system with the initial condition $\XX_0 = (-2, 4)$ can be generated by numerically integrating system (\ref{vdp_ode}) with time step $\Delta t = 0.01$ during time $T = 10$. Having this training data set, the proposed neural network can be fitted with the mean squared error (MSE) as a loss function based on the norm
$$
||\XX_{i+1} - \mathcal M\circ \XX_i ||=
||
\begin{pmatrix}
x_{i+1}\\
y_{i+1}\\
\end{pmatrix}
-W_0 - W_1
\begin{pmatrix}
x_i\\
y_i\\
\end{pmatrix}
-W_2
\begin{pmatrix}
x_i^2\\
x_iy_i\\
y_i^2
\end{pmatrix}
-W_3
\begin{pmatrix}
x_i^3\\
x_i^2y_i\\
x_iy_i^2\\
y_i^3
\end{pmatrix}
||.
$$
We implemented the above-described technique in Keras/TensorFlow and fitted a third-order Lie transform--based neural network with an Adamax optimizer.

\begin{figure}
\centering
\includegraphics[width=0.40\textwidth]{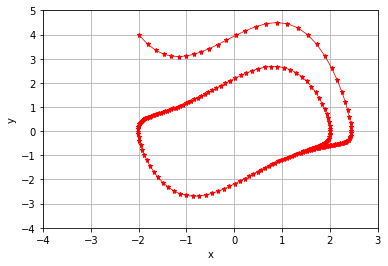}
\hskip 40pt
\includegraphics[width=0.40\textwidth]{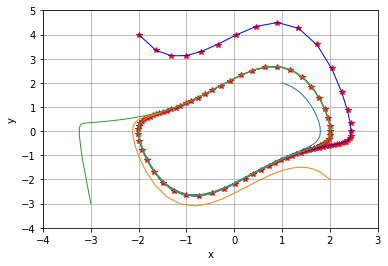}
\\
a) particular solution as the training set
\hskip 20pt
b) prediction for new initial conditions
\caption{Training data for the neural network (red dots) and provided predictions (lines).}
\label{fig3}
\end{figure}

\noindent
The generalization property of the network can be investigated by examining prediction not only from the training data set but also for new initial conditions. Fig. \ref{fig3} (a) shows the training set as a particular solution of the system with initial conditions $(-2, 4)$. Fig. \ref{fig3} (b) demonstrates predictions that are calculated starting at both the same initial condition and for the new points $(1, 2), (2, -2)$, and $(-3, -3)$. For the prediction starting from the training initial condition, the mean relative error of the predictions is $4.8\cdot10^{-5}$. For the new initial conditions, the mean error is $9.8\cdot 10^{-6}$.

\section{Partial Differential Equations}
\label{sec4}
Burgers' equation is a fundamental partial differential equation that occurs in various areas, such as fluid mechanics, nonlinear acoustics, gas dynamics, and traffic flow. This equation is also often used as a benchmark for numerical methods. For example, one of the problems proposed in the Airbus Quantum Computing Challenge \cite{ref19} is building a neural network that solves Burgers' equation with at least the same level of accuracy and higher computational performance as the traditional numerical methods. In the article \cite{ref20}, a feedforward neural network is trained to satisfy Burgers' equation and certain initial conditions, but the computational performance of the approach is not estimated. In this section, we demonstrate how to build a Lie transform--based neural network that solves Burgers' equation.
\subsection{The Finite Difference Method for Burgers' Equation}
Burgers' equation has a form
\begin{equation}
\label{burgers_de}
    \frac{\partial u(t,x)}{\partial t} + u(t,x)\frac{\partial u(t,x)}{\partial x} = \nu \frac{\partial^2 u(t,x)}{\partial x^2}.
\end{equation}

\noindent
Following the \cite{ref19} for benchmarking, we use an analytic solution
\begin{equation*}
    u(t,x) = -2\frac{\nu}{\phi(t,x)}\frac{d\phi}{dx}+4,\;\phi(t,x) = exp\frac{-(x-4t)^2}{4\nu(t+1)} + exp\frac{-(x-4t-2\pi)^2}{4\nu(t+1)},
\end{equation*}
and a traditional numerical method
\begin{equation}
\label{FDM}
    \frac{u^{n+1}_i - u^{n}_i}{\Delta t} + u^{n}_i\frac{u^{n}_i - u^{n}_{i-1}}{\Delta  x} = \nu \frac{u^n_{i+1} - 2u^n_i + u^n_{i-1}}{\Delta x^2},
\end{equation}
where $n$ stands for the time step, and $i$ stands for the grid node.

The equation (\ref{FDM}) presents a finite difference method (FDM) that consists of an Euler explicit time discretization scheme for the temporal derivatives, an upwind first-order scheme for the nonlinear term, and finally a centered second-order scheme for the diffusion term. The time step for benchmarking is fixed to $\Delta t = 2.5\cdot10^{-4}$ with the uniform spacing of $\Delta x = 2\pi/1000$. Thus, for the numerical solution for times from $t = 0$ to $t = 0.$5 on $x\in[0, 2\pi]$, the method requires the mesh with 1000 steps on space coordinate $x$ and 2000 time steps.

\begin{figure}
\centering
\includegraphics[width=0.40\textwidth]{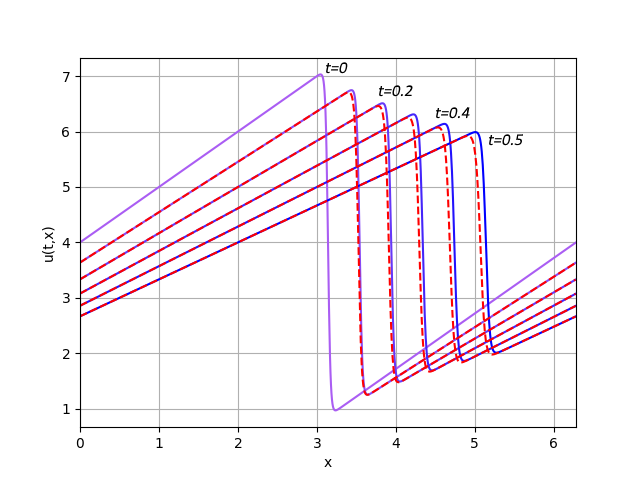}
\hskip 40pt
\includegraphics[width=0.40\textwidth]{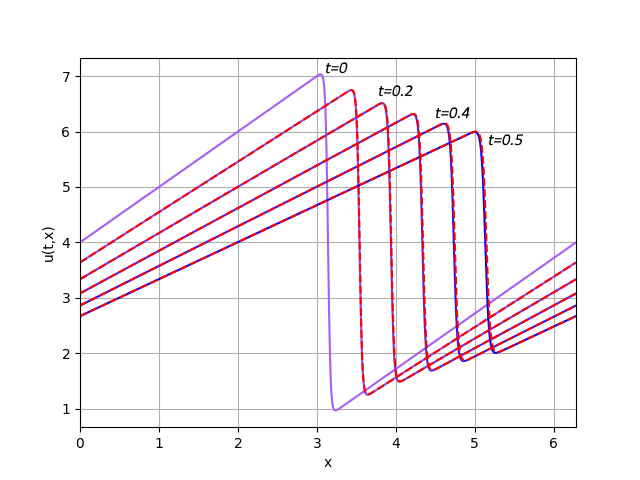}
\\
\hskip 20pt
a) FDM, $\Delta t = 2.5\cdot10^{-4}$
\hskip 60pt
b) matrix Lie map, $\Delta t = 1.25\cdot10^{-3}$
\caption{A benchmark (a) on mesh $1000\times 2000$ and Lie transform--based neural network (b) on mesh $1000\times500$.}
\label{fig4}
\end{figure}

It is indicated in \cite{ref19} that the FDM introduces a dispersion error in the solution (see Fig. \ref{fig4}, a). Such error can be reduced by increasing the mesh resolution, but then the time step should be decreased to respect the stability constraints of the numerical scheme.

\subsection{Lie Transform--Based Neural Network}
Though there are Lie group methods that directly apply Lie theory to PDEs \cite{ref21,ref22}, we utilize a different approach. We convert the equation (\ref{burgers_de}) to a system of ODEs and build a matrix Lie map in accordance with Section \ref{sec2} for this new system.

Assuming that the right-hand side of the equation (\ref{burgers_de}) can be approximated by a function $f(x, u(t,x))$ and considering this approximated equation as a hyperbolic one, it is possible to derive the system of ODEs

\begin{equation}
    \label{burgers_ode}
    \frac{d}{dt}
    \begin{pmatrix}
    \XX\\
    \UU
    \end{pmatrix}
    =
    \begin{pmatrix}
    \UU\\
    f(x, \UU)
    \end{pmatrix},
\end{equation}
where $\UU=(u_1, \ldots, u_{1000})$, $u_i(t) = u(t, x_i)$, and $\XX=(x_1, \ldots, x_{1000})$ is vector of discrete stamps on space. This transformation from PDE to ODE is well known and can be derived using the method of characteristics and direct method \cite{ref23}. If $f(x, u(t,x))$ is the same discretization as in (\ref{FDM}), then the equation (\ref{burgers_ode}) leads to the system of 2000 ODEs
\begin{equation*}
    \begin{aligned}
    x'_i &= u_i,\\
    u'_i &= f(\nu, u_{i+1}, u_i, u_{i-1}, x_{i+1}, x_i, x_{i-1}),
    \end{aligned}
\end{equation*}
which can be easily expanded to the Taylor series with respect to the $\XX$ and $\UU$ up to the necessary order of nonlinearity.

\begin{figure}
\centering
\includegraphics[width=0.45\textwidth]{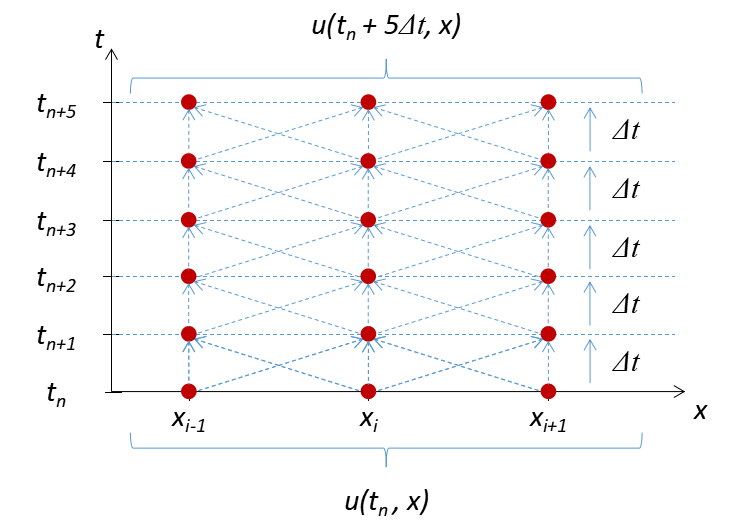}
\hskip 30pt
\includegraphics[width=0.45\textwidth]{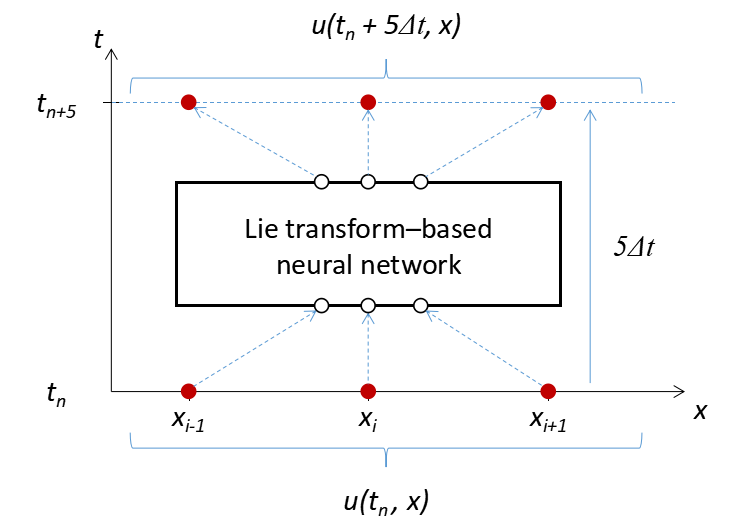}
\hskip 20pt
a) FDM, $\Delta t = 2.5\cdot10^{-4}$
\hskip 80pt
b) matrix Lie map, $\Delta t = 1.25\cdot10^{-3}$
\caption{Numerical schemes for FDM and Lie transform--based neural network for (\ref{burgers_de}).}
\label{fig5}
\end{figure}

Using this system of ODEs, we have built a Lie transform--based neural network for a time interval $\Delta t = 1.25\cdot10^{-3}$. This time step is five times larger than that used in the benchmarking (see Fig.~\ref{fig5}). The numerical solution provided by the neural network is presented in Fig. \ref{fig4} (b), and the accuracy and performance are compared in Table \ref{table2}.

\begin{table}
\centering
\caption{Comparison of simulation by FDM and matrix Lie map.}\label{table2}
\begin{tabular}{|l|c|c|c|c|}
\hline
\textbf{Method} & \textbf{Time} & \textbf{Mesh} & \textbf{Elapsed} & \textbf{MSE for} \\
 & \textbf{step} & \textbf{size} & \textbf{time} & $u(0.5,x)$ \\
\hline
FDM&\;\;\;\;$2.5\cdot10^{-4}$\;\;\;\;&\;\;\;\;$1000\times2000$\;\;\;\;&\;\;\;\;0.055 sec\;\;\;\;&\;\;\;\;$8.0\cdot10^{-2}$\;\;\;\;\\
\hline
Lie transform--based&&&&\\
neural network&$1.25\cdot10^{-3}$&$1000\times500$&0.016 sec&$5.5\cdot10^{-3}$\\
\hline
\end{tabular}
\end{table}

\noindent
The built polynomial neural network provides better accuracy with less computational time. If the FDM scheme is adjusted to a higher accuracy, the computational time will be increased even more. Accuracy is calculated as the MSE metric between the numerical solution and its analytic form at final time $t = 0.5$.

\section{Code}
The implementation of the  Lie transform--based neural network in Keras/Tensor\-Flow %(\href{https://github.com/andiva/DeepLieNet/blob/master/core/Lie.py}{LieLayer.py}) 
and the algorithm for map building for autonomous systems %(\href{https://github.com/andiva/DeepLieNet/blob/master/core/Lie_map_builder.py}{Lie\_map\_builder.py})
are provided at the GitHub repository: \href{https://github.com/andiva/DeepLieNet}{https://github.com/andiva/DeepLieNet}. 
The notebook 
\href{https://github.com/andiva/DeepLieNet/tree/master/demo/VanderPol.ipynb}{https://github.com/andiva/DeepLieNet/tree/master/demo/VanderPol.ipynb}\\
corresponds to Section \ref{sec3} and consists of simulation, the definition of metrics for accuracy estimation, neural network configuration, and fitting. The notebook 
\href{https://github.com/andiva/DeepLieNet/tree/master/demo/Burgers.ipynb}{https://github.com/andiva/DeepLieNet/tree/master/demo/Burgers.ipynb}
reproduces the results presented in Section \ref{sec4}.

\section{Conclusion}
In the article, we demonstrate the solving of differential equations with polynomial neural networks that are based on matrix Lie maps. Since the weights of the proposed neural network can be directly calculated from the equations, it does not require fitting with respect to the initial condition. Built at once, the neural network can be considered as a model of the system and can be used for simulation with different initial conditions.

In the case of large time steps for map calculating, the proposed approach can significantly outperform traditional numerical methods. For Burgers' equation, the computational performance is increased several times with the same level of accuracy. For some problems in the charged particle dynamics simulation, the performance is increased a thousand times with an appropriate accuracy in comparison with the traditional step-by-step integrating \cite{ref24,ref25}.

The proposed neural network can be used for data-driven identification of the systems. It may provide a high level of generalization when learning dynamical systems from data. As shown with the Van der Pol oscillator, learning the dynamics of the system with only a particular solution is possible. The neural network presented in Section \ref{sec4} can be additionally fitted to satisfy the initial conditions. In this sense, the training will provide an optimal numerical approach with respect to certain initial conditions.

The limitations of the data-driven approach for large-scale systems, optimal network configuration, and noisy data consideration should be examined in further research.


\end{document}